\documentclass{article}
\usepackage{spconf,amsmath,epsfig}
\usepackage{amssymb}
\usepackage{subfigure}
\usepackage{algorithm}
\usepackage{algorithmicx}
\usepackage{algpseudocode}
\usepackage{setspace}
\usepackage{longtable}
\usepackage{diagbox}
\usepackage{multirow}
\usepackage{color}
\usepackage[switch]{lineno}

\title{Imperceptible Adversarial Examples for Fake Image Detection}
\name{
Quanyu Liao$^1$, Yuezun Li$^3$, Xin Wang$^2{}^{\dagger}$\thanks{{\bf $\dagger$} Corresponding authors: Xin Wang (xinw@keyamedna.com), Xi Wu (xi.wu@cuit.edu.cn).}, Bin Kong$^2$, \textit{Bin Zhu}$^4$, Siwei Lyu$^3$, \textit{Youbing Yin}$^2$, \textit{Qi Song}$^2$, \textit{Xi Wu}$^1{}^{\dagger}$
}

\address{
$^1$ Chengdu University of Information Technology, Chengdu, China \\
$^2$ Keya Medical, Seattle, USA\\
$^3$ University at Buffalo, State University of New York, USA \\
$^4$ Microsoft Research Asia, Beijing, China 
}

\begin{document}
%
\maketitle
\begin{abstract}
Fooling people with highly realistic fake images generated with Deepfake or GANs brings a great social disturbance to our society. Many methods have been proposed to detect fake images, but they are vulnerable to adversarial perturbations -- intentionally designed noises that can lead to the wrong prediction. Existing methods of attacking fake image detectors usually generate adversarial perturbations to perturb almost the entire image. This is redundant and increases the perceptibility of perturbations. In this paper, we propose a novel method to disrupt the fake image detection by determining key pixels to a fake image detector and attacking only the key pixels, which results in the $L_0$ and the $L_2$ norms of adversarial perturbations much less than those of existing works. Experiments on two public datasets with three fake image detectors indicate that our proposed method achieves state-of-the-art performance in both white-box and black-box attacks.
\end{abstract}

\section{Introduction}
\label{sec: introduction}
   
The development of generative adversarial networks~(GANs) \cite{Goodfellow2014Generative} enables generating high-quality images. 
Different GANs have been proposed to fit different applications. One of these applications is DeepFake~\cite{deepfake}, which can swap the face of a source subject with that of a target subject while retaining original facial expressions with high realism. 
It can be easily abused for malicious purposes, such as replacing the original face in a pornographic video with the victim's face, which can cause a significant social disturbance.
   
Many methods~\cite{wang2020cnn, rossler2019faceforensics++} have been proposed to detect fake images and videos. These methods are usually based on CNNs and trained on public DeepFake datasets such as UADFV~\cite{yang2019exposing}, CelebDF-v2~\cite{li2020celeb}, and Faceforensics++~\cite{rossler2019faceforensics++}. Despite they can achieve high performance, these methods are vulnerable to adversarial perturbations \cite{goodfellow2014explaining, dong2018boosting}, which are intentionally designed noises that can fool these methods.

Several anti-forensics methods \cite{carlini2020evading, Gandhi2020Adversarial} have been proposed recently to evade fake detection using adversarial perturbations. They are based on either minimizing the distortion of perturbations on the entire image \cite{goodfellow2014explaining, carlini2017towards} or reducing the size of the attacking area without considering the distortion and the time consumption \cite{madry2017towards}. More specifically, gradient-based attack methods~\cite{Gandhi2020Adversarial} and $L_2$-distortion minimizing attack~\cite{carlini2020evading} change almost every pixel, which is unnecessary and makes resulting adversarial perturbations more perceptible.
Other adversarial methods~\cite{madry2017towards, carlini2020evading, modas2019sparsefool, 2017One, 0Sparse} aim to reduce the $L_0$ norm of generated perturbations. They have high time consumption, and their generated perturbations have a higher $L_2$ norm than that of perturbations generated by gradient-based attack methods or $L_2$-distortion minimizing attacks.

\begin{figure}[t]
    \centering
    \includegraphics[width=1\linewidth]{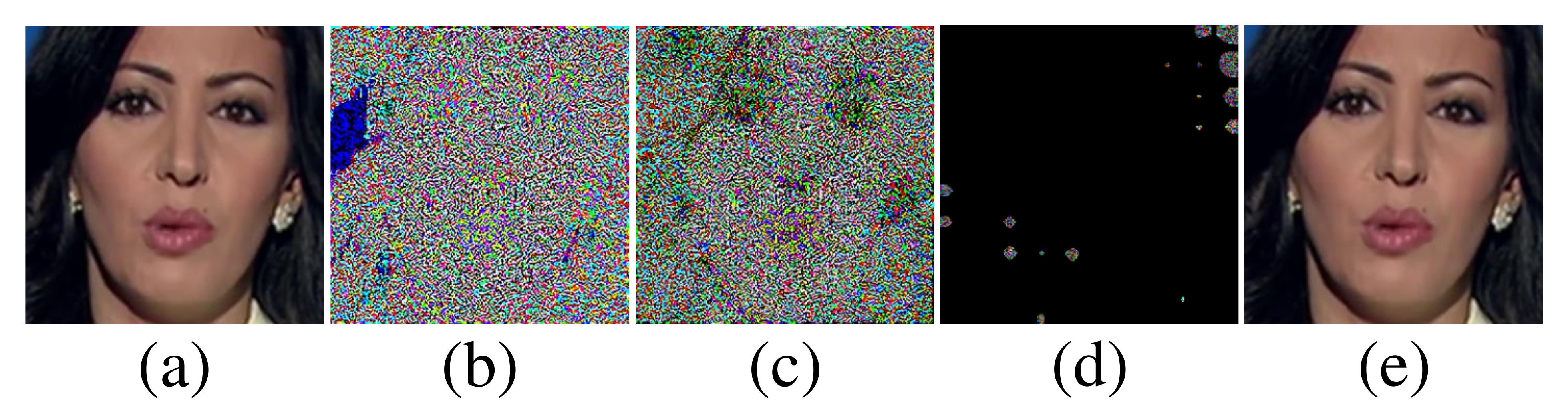}
    \vspace{-0.75cm}
    \caption{\small Comparison of generated adversarial perturbations of FGSM, DeepFool, and our Key Region Attack~(KRA), for fake image detection. (a) is a fake image generated by DeepFake and detected as a fake image by the detector~\cite{rossler2019faceforensics++}. (b) is an adversarial perturbation generated by FGSM~\cite{goodfellow2014explaining}. (c) is an adversarial perturbation generated by DeepFool~\cite{he2016deep}. (d) is an adversarial perturbation generated by our method KRA. (e) is an adversarial example generated by adding perturbation (d) on image (a), which is detected as a real image by the detector \cite{rossler2019faceforensics++}. (b)(c)(d) are obtained after normalizing the pixel intensity of perturbations to $[0, 255]$).}
    \vspace{-0.5cm}
\end{figure}

To address the aforementioned limitations of existing adversarial attacks, we propose in this paper a novel method, called \emph{Key Region Attack}~(KRA), to attack only a small portion of an input image while minimizing the distortion of added perturbations. Existing methods~\cite{Gandhi2020Adversarial, carlini2020evading} minimize either $L_0$ or $L_2$ norm of perturbations. They cannot make perturbations both sparse and imperceptible. To make perturbations sparse and imperceptible, KRA aims to minimize both $L_2$ and $L_0$ norms of perturbations simultaneously. More specifically, KRA leverages the gradient of multiple convolutional layers and spatial information to efficiently extract key pixels that fake detection relies on, and then apply a conventional adversarial method to generate adversarial perturbations by modifying only the key pixels. 
Experimental results show that KRA achieves state-of-the-art attack performance with high efficiency and significantly reduce both $L_0$ and $L_2$ norms of generated disturbances. 

Our method has the following key features, which are also our main contributions:
\textbf{1)} We propose to attack only pixels key to fake detection in an image in generating adversarial disturbances. Attacking key pixels alone can greatly disrupts fake detection while  minimizing both $L_0$ and $L_2$ of perturbations at the same time.
\textbf{2)} Instead of using the heatmap of the last layer as key pixels, which leads to a large $L_0$ norm,  we propose Multi-Layers Key Semantic Region Selection~(MLSKRS) to combine heatmaps from both shallow and deep layers to extract key pixels in an image. The resulting set of key pixels is much smaller yet effective. 
\textbf{3)} KRA is designed as a container to support integrating different adversarial attack methods, include FGSM, Deepfool, etc.

\begin{figure}[t]
    \centering
    \includegraphics[width=0.9\linewidth]{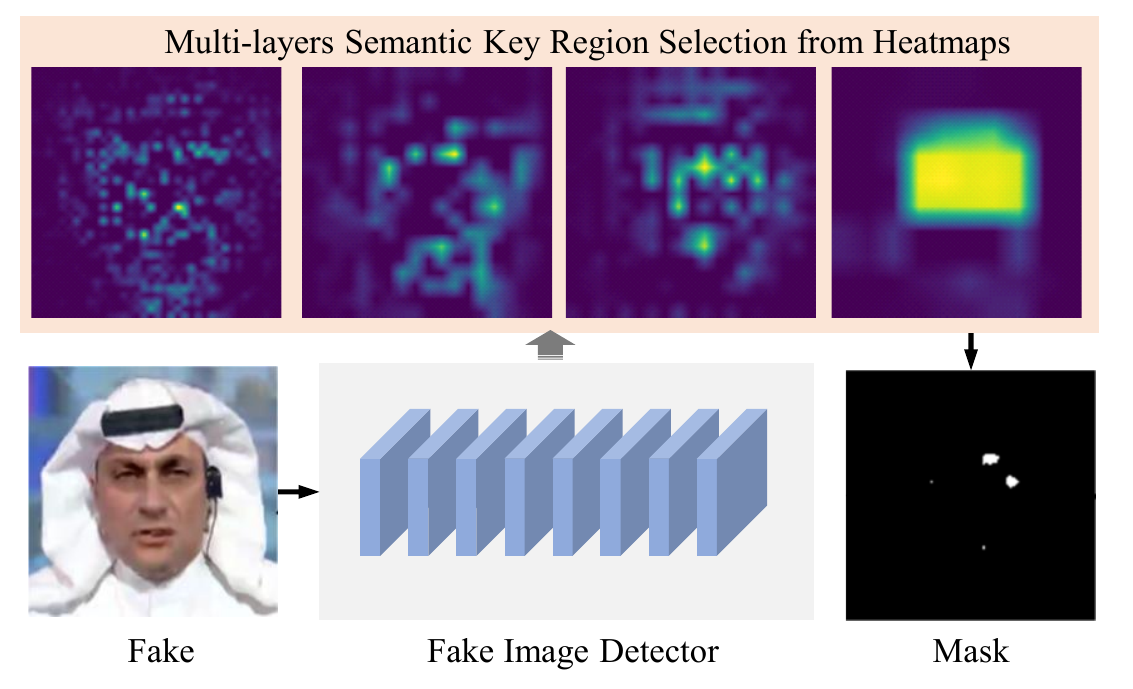}
    \vspace{-0.25cm}
    \caption{\small Illustration of Multi-layers SemanticKey Region Selection~(MLSKRS): MLSKRS extracts mask $m_i$ for each convolutional layer $l_i$ and combines masks $m_i$ to generates a final mask $\cal M$.}
    \label{overall_fig}
    \vspace{-0.5cm}
\end{figure}

\section{KRA Method}
\label{sec: method}

\subsection{Problem Definition}


As mentioned before, we aim to minimize both $L_2$ and $L_0$ norms of perturbations simultaneously to generate sparse and imperceptible adversarial disturbances of fake detectors. This can be formulated as follows:

\begin{equation}
    \begin{aligned}
        \mathop{minimize} \quad & {\Vert r \Vert}_{2} + {\Vert r \Vert}_{0}\\
        subject \quad  & f(x+r) \not= f(x)\\
    \end{aligned}
\label{problem_defination}
\end{equation}
where $x$ is a clean input image, $f(x)$ is the predicted result of the fake detector, and $r$ is an adversarial perturbation.


Minimizing the $L_0$ norm of perturbations is an $NP$ problem~\cite{modas2019sparsefool}. To address this problem, KRA does not minimize the $L_0$ norm directly. Instead, it finds key pixel regions of the input image that fake detection relies on and generates adversarial perturbations by modifying only the found key pixels. In this way, the $L_0$ norm of perturbations can be efficiently reduced, and resulting adversarial examples are imperceptible. We reformulate the problem as follows:


\begin{equation}
    \begin{aligned}
        \mathop{minimize} \quad & ||r||_{2} + ||{\cal M}||_{0} \\
        subject to \quad  & f(x+(r \cdot {\cal M})) \not= f(x)\\
    \end{aligned}
\label{problem_defination_kra}
\end{equation}

\noindent where $\cal M$ is the set of key pixels of the input image.


\subsection{Multi-layers Semantic Key Region Selection}
\label{sec_mlskrs}


Convolutional layers retain spatial information~\cite{Zhou2016Learning, selvaraju2017grad}, which can indicate which regions of an input image have a higher response to the output of the classifier. This information is exploited in KRA to extract key pixels to fake detection. Only key pixels are disturbed.  

A straightforward method is to generate a heatmap based only on the last convolutional layer as in prior work~\cite{selvaraju2017grad} (see the top right heatmap in Fig.~\ref{overall_fig}). This method does not suppress the $L_0$ norm of adversarial perturbations well since deeper convolutional layers correspond to a larger receptive field, leading to a large and continuous region of high response in the generated heatmap. To address this problem, we propose Multi-layers Semantic Key Region Selection~(MLSKRS) to generate heatmaps based on multiple convolution layers (see the heatmaps in Fig.~\ref{overall_fig}), including both shallow layers and deep layers. Since shallow layers correspond to a small receptive field, MLSKRS can significantly reduce the $L_0$ norm of the heatmap while maintaining effective gradients generated on deep layers.

\renewcommand{\algorithmicrequire}{\textbf{Input:}}
\renewcommand{\algorithmicensure}{\textbf{Output:}}
\begin{algorithm}[t]
    \setstretch{1.3}
    \caption{Multi-layers Semantic Information Key Region Selection~(MLSKRS)}
    \label{mlskrs}
    \begin{algorithmic}[1]
        \Require
        image $\mathop{x}$, attack threshold $\mathop{t_\alpha}$,
        fake image detector $\mathop{f(\cdot)}$, set of target layers $\mathop{\cal L}$, 
        operation of upsampling $\mathop{\Upsilon}$
        \Ensure
        final mask $\mathop{\cal M}$
        \For{$l_i : \cal L$}
            \For{$h^j_i : l_i$}
                \State$\mathop{g^j_i = \frac{\partial f(x)}{\partial h^j_i}}$
                
            \EndFor
            \State$\mathop{G_i = \sum_{j=1}^k g^j_i}$
            \State$\mathop{\hat{G}_i = \Upsilon(\frac{G_i - \text{min}(G_i)}{\text{max}(G_i) - \text{min}(G_i)})}$
            \State$\mathop{m_i = \delta (\hat{G}_i > t)}$
        \EndFor
        \State$\mathop{{\cal M} = \prod_{i=1}^n  m_i}$
        \State{return $\mathop{{\cal M}}$}
    \end{algorithmic}
\end{algorithm}

Denote ${\cal L} = \{l_i\}^n_{i=1}$ as the set of layers we consider and $l_i = \{h^j_i\}^k_{j=1}$ as the $i$-th layer that contains $k$ channels. 
For layer $l_i$, MLSKRA first calculates the gradient of channel $h^j_i$ and then sums up the gradients of all channels as
\begin{equation}
    \begin{aligned}
        g^j_i & = \frac{\partial f(x)}{\partial h^j_i}, ~~ G_i = \sum_{j=1}^k g^j_i \\
    \end{aligned}
\label{mlskrs-2}
\end{equation}
then gradient $G_i$ is normalized to $[0, 1]$ and upsampled to the size of image $x$: 
\begin{equation}
    \begin{aligned}
        \hat{G}_i & = \Upsilon(\frac{G_i - \text{min}(G_i)}{\text{max}(G_i) - \text{min}(G_i)})
    \end{aligned}
\label{mlskrs-3}
\end{equation}
where $\Upsilon(\cdot)$ denotes the operation of upsampling. The mask corresponding to layer $l_i$ can be obtained using an attack threshold $t$: 
\begin{equation}
    \begin{aligned}
        m_i = \delta (\hat{G}_i > t)
    \end{aligned}
\label{mlskrs-4}
\end{equation}
where $\delta$ is an impulse response that turns a pixel greater than $t_\alpha$ to $1$, otherwise to $0$. MLSKRS combines all masks as
\begin{equation}
    \begin{aligned}
        {\cal M} & = \prod_{i=1}^n  m_i
    \end{aligned}
\label{mlskrs-5}
\end{equation}
where $\cal M$ denotes the final mask. The procedure of MLSKRS is summarized in Alg.~\ref{mlskrs}.

\renewcommand{\algorithmicrequire}{\textbf{Input:}}
\renewcommand{\algorithmicensure}{\textbf{Output:}}
\begin{algorithm}[t]
    \setstretch{1.3}
    \caption{Key Region Attack~(KRA)}
    \label{kra}
    \begin{algorithmic}[1]

        \Require
        image $\mathop{x}$, fake detector $\mathop{f}$, attack method $\phi$,
        initial attack threshold $\mathop{t_{\alpha}}$, lowest attack threshold $\mathop{t'}$,
        reduction step size of threshold $\beta$
        \Ensure
        perturbation $\mathop{r'}$

        \State{Initialize: $\mathop{u = 0, t_0 = t_{\alpha}, r'_{-1} \in[0]^{W\times H}}$}

        \While{ \text{True} }            
            \State$\mathop{{\cal M}_{u} = \text{MLSKRS}(x, t_u)}$
            \State$\mathop{r_u = \phi(x+r'_{u-1})}$
            \State$\mathop{r'_{u} = r'_{u-1} + r_u \cdot {\cal M}_u}$
            \If{$\mathop{f(x + r'_{u}) \neq f(x)}$} 
            \State {\text{break}} 
            \EndIf
            \State$t_{u+1} = \max(t_u - \beta, t')$
            \State$\mathop{u = u + 1}$
            
        \EndWhile

        \State{return $r'_{u}$}
        
    \end{algorithmic}
\end{algorithm}

\subsection{Key Region Attack}
\label{sec_kra}

Key Region Attack~(KRA) attacks the key regions obtained from MSKRS iteratively. At each iteration, KRA dynamically updates the attack threshold and calculates key regions using MSKRS, and then utilizes an adversarial attack method to attack the key regions. Various state-of-the-art adversarial attack methods can be used, e.g., PGD \cite{carlini2017towards}, DeepFool \cite{moosavi2016deepfool}), etc. As a result, KRA is a container that can flexibly integrate various attack methods to meet different requirements.

Let the selected adversarial attack method be $\phi$ and $u$ be the iteration index. The threshold is initialized as $t_0 = t_{\alpha}$, where $t_{\alpha}$ is a constant. At each iteration, the key region mask is obtained: ${\cal M}_u = \text{MLSKRS}(x, t_u)$. Attacking method $\phi$ is employed to generate an adversarial perturbation $r_{u}$: $r_{u} = \phi(x + r'_{u-1})$. Then $r_{u}$ is masked by ${\cal M}_u$ and then added with $r'_{u-1}$ to generate $r'_u$. Finally, $x+r'_u$ is sent to detector $f$ to see whether this attack succeeds. If the attack succeeds, the iteration is terminated. Otherwise the threshold is updated as $t_{u+1} = \max(t_u - \beta, t')$ to gradually enlarge key regions, where $\beta$ is the reduction step size, and $t'$ is the lower bound of threshold. The procedure of KRA is shown in Alg.~\ref{kra}.

\begin{table*}[h]
    
    \begin{center}
    \resizebox{1.7\columnwidth}{!}{
    \begin{tabular}{|c|c|c|c|c|c|c|c|c|}
        \hline
        Attack Method      &DataSet            &Network        & Acc (Clean)    & Acc (Attack)   & ASR   &$P_{L_2}$   &$P_{L_0}$   &Time (s)    \\ \hline
        $L_2-attack$~\cite{carlini2020evading}   &Cnn-Synth-all      &Resnet-50      & 0.83          & 0.001         & 0.99  &$1\times10^{-1}$ &---        &---        \\ \hline
        $L_0-attack$~\cite{carlini2020evading}   &Cnn-Synth-Fake     &Resnet-50      & 0.83          & 0.010         & 0.99  &---        &11\%       &---        \\ \hline
        $L_0-attack$~\cite{carlini2020evading}   &Cnn-Synth-Real     &Resnet-50      & 0.83          & 0.021         & 0.75  &---        &11\%       &---        \\ \hline
        KRA-PGD         &Cnn-Synth-all      &Resnet-50      & 0.83          & \textbf{0.001}         & \textbf{0.99}  &$\mathbf{4\times10^{-4}}$     &\textbf{0.1\%}      &\textbf{0.13}       \\ \hline 
        KRA-PGD         &FaceForensics-all  &Xception       & 0.99          & 0.006         & \textbf{0.99}  &$8\times10^{-4}$     &\textbf{0.9\%}      &\textbf{0.15}       \\ \hline
        KRA-PGD        &FaceForensics-all  &Resnet-50      & 0.99          & 0.001         & 0.99  &$1.5\times10^{-3}$     &0.9\%      &0.22       \\ \hline
        KRA-PGD         &FaceForensics-all  &Resnet-101     & 0.99          & 0.001         & 0.99  &$1.7\times10^{-3}$     &1\%        &0.4        \\ \hline
        KRA-DeepFool &FaceForensics-all  &Xception     & 0.99          & \textbf{0.001}         & 0.99  &$\mathbf{6\times10^{-5}}$     &21\%        &2        \\ \hline
    \end{tabular}}
    \end{center}
    \vspace{-0.5cm}
    \caption{
    \small { Results of White-Box Attack. Acc (Clean) means the accuracy obtained from clean inputs. Acc (Attack) denotes the accuracy obtained from adversarial examples. The 'Time' column shows the average attack time. FaceForensics-all and Cnn-Synth-all mean that both real images and fake images are used in the attack. Cnn-Synth-Fake means attacking only fake images, while Cnn-Synth-Real means attacking only real image.}
    }
    \label{white-box}
    \vspace{-0.5cm}   
\end{table*}

\begin{table}[t]
\begin{center}
\resizebox{0.95\columnwidth}{!}{
\begin{tabular}{|c|c|c|c|c|c|c|c|c|}
\hline
            & \multicolumn{2}{c|}{Xception} & \multicolumn{2}{c|}{Resnet-50} & \multicolumn{2}{c|}{Resnet-101} & \multicolumn{2}{c|}{Inception-v3} \\ \hline
            & Acc            & ATR          & Acc            & ATR           & Acc             & ATR           & Acc             & ATR             \\ \hline
Clean      & 0.99           & ---          & 0.99           & ---           & 0.99            & ---           & 0.99            & ---             \\ \hline
Xception  & 0.006          & 1.00         & 0.43           & 0.56          & 0.51            & 0.49          & 0.36            & 0.63            \\ \hline
Resnet-50  & 0.49           & 0.50         & 0.001          & 1.00          & 0.46            & 0.53          & 0.56            & 0.43            \\ \hline
Resnet-101 & 0.48           & 0.51         & 0.42           & 0.57          & 0.001           & 1.00          & 0.59            & 0.40            \\ \hline
\end{tabular}}
\end{center}
\vspace{-0.5cm}
\caption{\small Results of black-box attack using KRA-PGD. 'Clean' means the accuracy obtained from clean inputs. The first column denotes the original detector that adversarial perturbations are generated with. The first row denotes the target detector that the adversarial examples are used to attack.}
\label{black-box}
\vspace{-0.75cm}
\end{table}

\section{Experimental Evaluation}
\label{sec: experiment}



\noindent\textbf{Datasets.} We validate attack methods on two public datasets, FaceForensics++ and CNN-Synthesis. \textit{FaceForensics++~\cite{rossler2019faceforensics++}} includes 1000 real videos and 1000 fake videos generated by Deepfake~\cite{deepfake}.
\textit{CNN-Synthesis~\cite{wang2020cnn}} is generated with a variety of GANs on real images collected from the Internet.

\noindent\textbf{Fake Image Detectors.}
\textit{CNN-Synthesis Detector}~\cite{wang2020cnn} is based on Resnet50 \cite{he2016deep} and trained on CNN-Synthesis.
\textit{Xception~\cite{rossler2019faceforensics++}} is based on XceptionNet and trained on FaceForensics++. We also train three additional fake image detectors, Inceptionv3~\cite{szegedy2016rethinking}, Resnet50, and Resnet101~\cite{he2016deep}, as attacking targets.

\noindent\textbf{Implementation Details.}
Our method is implemented using Pytorch 1.1.0. The experiments were run on Ubuntu 18.04 with one GPU of Nvidia Tesla T4 and the following values of the parameters in our method: 
$t_{\alpha} = 0.8$, $t' = 0.1$, $\beta = 0.1$.

\noindent\textbf{Evaluation Metrics.} We use three metrics to evaluate the attacking performance of generated adversarial examples:

\noindent 
\textit{(1) Attack Success Rate (ASR)}
\label{ASR} to measures the attack success rate of adversarial examples:
$ASR = 1-({acc_{attack}}/{acc_{clean}})$, where $acc_{attack}$ is the accuracy of the detector with adversarial examples, $acc_{clean}$ is the accuracy of clean inputs.

\noindent
\textit{(2) Attack Transfer Ratio (ATR)} to measure the black-box performance: $ATR = {ASR_{target}}/{ASR_{origin}}$, where $\mathop{ASR_{target}}$ represents the $\mathop{ASR}$ of attacking the target detector, and $\mathop{ASR_{origin}}$ denotes the $\mathop{ASR}$ of attacking the detector with which the adversarial examples are generated from.

\noindent
\textit{(3) Perceptibility.} We use both $L_0$ and $L_2$ norms of perturbations to measure perceptibility, denoted as $P_{L_0}$ and $P_{L_2}$, respectively. A lower value indicates more imperceptible. 

\noindent\textbf{White-Box Attack.} The attack performance of KRA is evaluated using PGD \cite{carlini2017towards},  denotes as KRA-PGD, and Deepfool \cite{moosavi2016deepfool}, denoted as KRA-Deepfool. The white-box attack results are summary in Table~\ref{white-box}. KRA achieves the state-of-the-art white-box attack performance: the accuracy of the three fake detectors is reduced to lower than $0.01$ under KRA's attack. Nearly all images of FaceForensics++ and CNN-Synthesis have been attacked successfully, including both attacking real images into the fake or attacking fake images into the real. Compared with the $L_0$-attack and $L_2$-attack \cite{carlini2020evading}, KRA has significantly reduced both $L_0$ norm and $L_2$ norm of adversarial perturbations, indicating that KRA generates much more imperceptible adversarial perturbations. We can see from Table~\ref{white-box} that the $L_2$ norm of the $L_2$-attack is nearly $100$ times higher than that of KRA, and the $L_0$ norm of the $L_0$-attack is $10$ times higher. We can also see that the $L_2$ norm of adversarial perturbations generated with KRA using Deepfool is reduced $10$ times as compared with that of KRA using PGD.

\begin{figure}[t]
    \centering
    \includegraphics[width=1.0\linewidth]{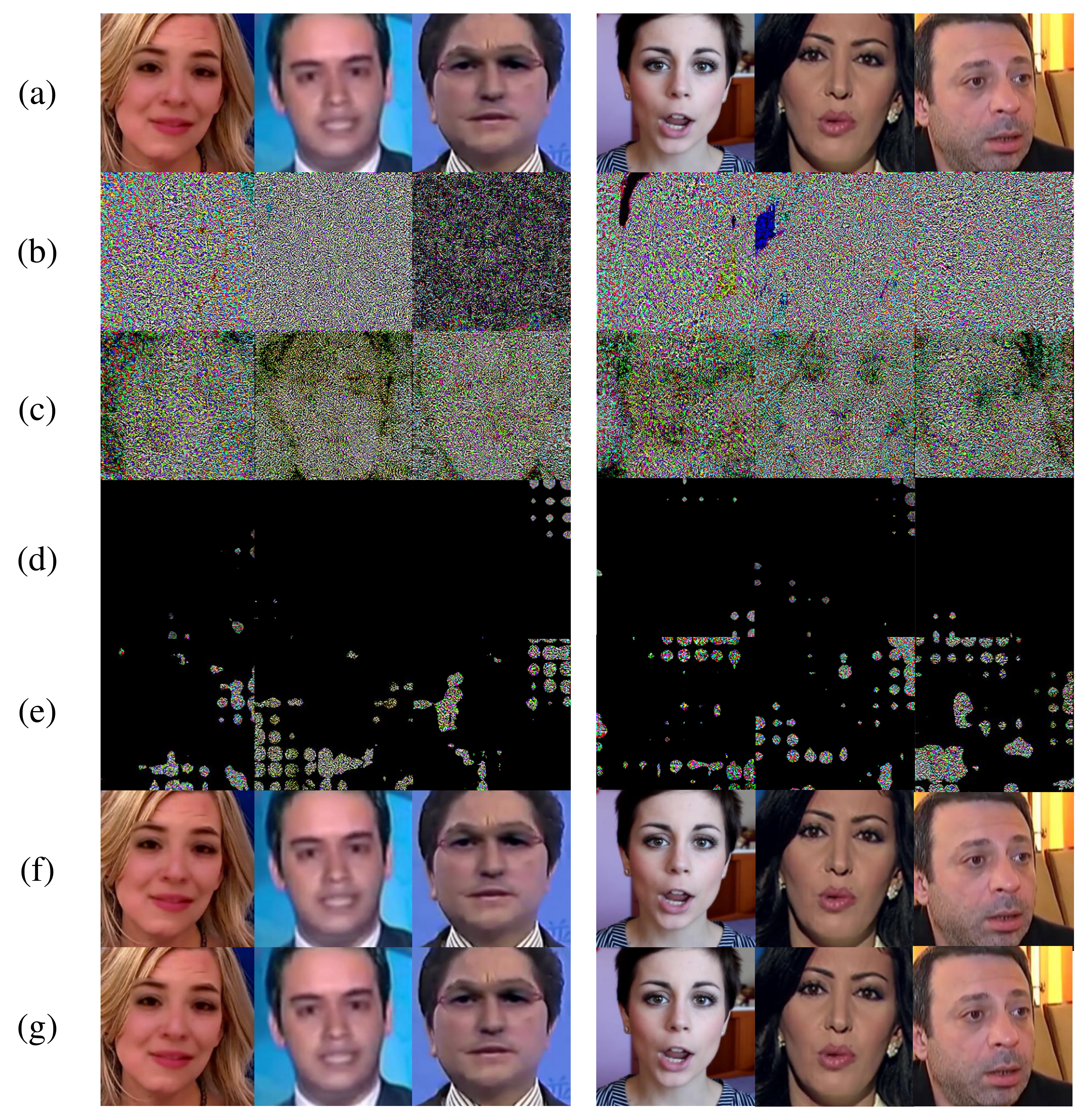}
    \vspace{-0.5cm}
    \caption{\small Qualitative results of KRA using PGD and Deepfool on FaceForensics~\cite{rossler2019faceforensics++}. The left three columns are fake images, the right three columns are real images. (a) are clean images without any perturbation. (b) are generated by PGD. (c) are generated by Deepfool. (d) are generated by KRA-PGD. (e) are generated by KRA-Deepfool. (f) and (g) are adversarial examples generated by KRA-PGD and KRA-Deepfool, respectively. (b),(c),(d) and (e) are obtained by normalizing the pixel intensity of perturbations to $[0, 255]$.}
    \label{qualitative}
    \vspace{-0.5cm}
\end{figure}

\noindent\textbf{Black-Box Attack.}  In this evaluation, adversarial examples are generated from a detector, called \emph{original detector}, and used to attack another detector, called \emph{target detector}. Black-box attack results are summarized in Table~\ref{black-box}. We can see from the table that ATRs of KRA on different detectors are all within $50\%\pm15\%$. Adversarial examples generated from Xception have a higher ASR on Inception-v3 than that on Resnet. Adversarial examples generated from the one Resnet have a high ASR on attacking the other Resnet, which can be explained that the two Resnet detectors learn relatively similar features due to their similar structures.

Both black-box and white-box attacks have the same setting. That means the $P_{L_0}$ and $P_{L_2}$ of perturbations in the black-box attack are identical to those in the white-box attack. Table~\ref{black-box} shows that KRA has a good black-box attack performance although adversarial disturbances generated by KRA have low $L_2$ norm and $L_0$ norm.
The black-box attack performance of KRA can be further improved by loosening the $L_2$ and $L_0$ restriction at the cost of worsen perceptibility or using more aggressive attack methods.

\noindent\textbf{Qualitative Analysis.} Fig.~\ref{qualitative} shows qualitative examples generated by KRA using PGD and Deepfool as the integrated attack method. We can see that perturbations appear only on key-regions, and a generated adversarial perturbation is hard for humans to distinguish.

\section{Conclusion}

In this paper, we have presented the Key Region Attack~(KRA) that generates imperceptible adversarial examples for attacking fake image detectors. 
KRA can flexibly integrate various attack methods to meet different requirements. Compare with previous methods, our method achieves the state-of-the-art performance for both white-box and black-box attacks, and adversarial examples generated with our methods are more imperceptible than those with previous methods.


\end{document}